\documentclass[10pt,twocolumn,letterpaper]{article}
\usepackage[pagenumbers]{cvpr}%

\emergencystretch=\maxdimen
\usepackage[dvipsnames]{xcolor}
\usepackage{array}

\usepackage{svg}
\usepackage{lineno}
\usepackage{listings}
\usepackage{longtable}
\usepackage{multirow}

\definecolor{cvprblue}{rgb}{0.21,0.49,0.74}
\usepackage[pagebackref,breaklinks,colorlinks,allcolors=cvprblue]{hyperref}

\title{SoccerChat: \\Integrating Multimodal Data for Enhanced Soccer Game Understanding}

\author{
Sushant Gautam\\
SimulaMet and OsloMet, Norway\\
{\tt\small sushant@simula.no}
\and
Cise Midoglu\\
Forzasys, Norway\\
{\tt\small cisemidoglu@gmail.com}
\and
Vajira Thambawita\\
SimulaMet, Norway\\
{\tt\small vajira@simula.no}
\and
Michael A. Riegler\\
SimulaMet, Norway\\
{\tt\small michael@simula.no}
\and
Pål Halvorsen\\
SimulaMet, OsloMet, and Forzasys, Norway\\
{\tt\small paalh@simula.no}
\and
Mubarak Shah\\
University of Central Florida, USA\\
{\tt\small shah@crcv.ucf.edu}
}

\begin{document} 
\maketitle

\begin{abstract}
The integration of artificial intelligence in sports analytics has transformed soccer video understanding, enabling real-time, automated insights into complex game dynamics. Traditional approaches rely on isolated data streams, limiting their effectiveness in capturing the full context of a match. To address this, we introduce SoccerChat, a multimodal conversational AI framework that integrates visual, and textual data for enhanced soccer video comprehension. Leveraging the extensive SoccerNet dataset, enriched with jersey color annotations and automatic speech recognition (ASR) transcripts, SoccerChat is fine-tuned on a structured video instruction dataset to facilitate accurate game understanding, event classification, and referee decision making. We benchmark SoccerChat on action classification and referee decision-making tasks, demonstrating its performance in general soccer event comprehension while maintaining competitive accuracy in referee decision making. Our findings highlight the importance of multimodal integration in advancing soccer analytics, paving the way for more interactive and explainable AI-driven sports analysis.
\end{abstract}
\\
\url{https://github.com/simula/SoccerChat}

\section{Introduction}
\label{sec:intro}

The integration of artificial intelligence (AI) into sports analytics has transformed our understanding of complex game dynamics, particularly in soccer. As the world's most popular sport, soccer generates vast data through matches, which, when effectively analyzed, provides valuable insights into team strategies, player performances, and game mechanics. Traditional analysis relies on manual annotation and isolated data streams, limiting depth and efficiency. The rise of multimodal AI, combining visual, auditory, and textual data, offers a holistic approach to soccer video understanding, enabling automated, real-time analysis that mirrors human-like comprehension.

Despite significant advancements, several challenges persist in the realm of soccer video analysis. One primary issue is the scarcity of comprehensive, annotated datasets that encompass the multifaceted nature of soccer matches. Existing datasets often lack synchronization between modalities or fail to capture the contextual nuances essential for accurate interpretation. Moreover, current models may struggle with the dynamic and unstructured environment of soccer games, where rapid movements, frequent occlusions, and complex interactions are prevalent. This complexity necessitates sophisticated models capable of integrating and processing multimodal data to achieve a coherent understanding of the game.

In response to these challenges, we introduce \textbf{SoccerChat}, a multimodal conversational AI framework designed specifically for comprehensive soccer video understanding. SoccerChat leverages the extensive SoccerNet dataset, enhancing it with additional annotations such as jersey color identification and automatic speech recognition (ASR) transcripts. This enriched dataset serves as the foundation for training our model to interpret complex game scenarios and generate contextually relevant responses. By integrating visual, auditory, and textual data, SoccerChat aims to emulate a human-like understanding of soccer matches, facilitating real-time analysis and interactive applications.

This paper presents the following key contributions:

\begin{itemize}
    \item \textbf{Dataset Enhancement:} We have curated a unique video instruction dataset with 49,120 question-answer pairs derived from SoccerNet videos, enriched with jersey color annotations and ASR transcripts, providing a valuable resource for training and evaluating soccer analysis models.
    \item \textbf{Model Development:} We introduce \textbf{SoccerChat}, a conversational AI model fine-tuned on the enhanced dataset, capable of understanding and generating contextually appropriate responses based on the multimodal inputs.
    \item \textbf{Comprehensive Evaluation:} Through different experiments, we demonstrate SoccerChat's proficiency in tasks such as event classification, and referee decision making.
\end{itemize}

To encourage reproducibility and further research, we release all artifacts, including the SoccerChat dataset, fine-tuned model weights for all variants, evaluation code, and intermediate files (LLM generations and scores). %

The remainder of this paper is structured as follows: Section 2 reviews related work on soccer video analysis and multimodal datasets. Section 3 details the methodology, including dataset construction, model configurations, and integration techniques. Section 4 presents the experimental setup and evaluation protocols, followed by the results in Section 5 and discussions in Section 6. Finally, Section 7 concludes the paper and sets the stage for future research.

\section{Related Work}
\label{sec:related_work}

\subsection{Soccer Video Understanding}
Soccer video understanding has gained significant traction in recent years, driven by advancements in multimodal data integration~\cite{Gautam2023OctBridging,SoccerStorytelling, matchvision2024}. This field focuses on analyzing player movements, game dynamics, and contextual events in soccer matches~\cite{Survey_soccer_2022}. Foundational tasks such as player segmentation, detection, tracking, re-identification, and keypoint detection are critical for understanding player interactions and strategies~\cite{SoccerNetTracking, SoccerNetV2DenseVideo}. Beyond player analysis, key research areas include action recognition, summarization, commentary generation, and replay grounding~\cite{SoccerNetV1, SoccerNetV2, SoccerNetV2DenseVideo,SoccerNetEchoes,SoccerSummarization,SoccerStorytelling}. The evolution of soccer video understanding also involves camera calibration, sports health monitoring, intelligent refereeing, and tactical analysis~\cite{SoccerNetCameraCalib, soccerathletehealth, SoccerNetXVARS, tacticalperformanceanalysis}, all of which benefit from challenges like \textit{SoccerNet}~\cite{SoccerNet2023}. Recent developments in Vision-Language Models (VLMs) and Multimodal Large Language Models (MLLMs) have further propelled progress, enabling comprehensive, real-time, and context-aware analysis of soccer videos~\cite{Gautam2023OctBridging, sportu2024}.

\subsection{Datasets for Soccer Video Analysis}

The advancement of soccer video understanding relies heavily on robust datasets~\cite{SoccerNetV1,Survey_soccer_2022}. One seminal contribution is the SoccerNet dataset comprising 500 complete soccer matches from major European leagues, totaling 764 hours of footage~\cite{SoccerNetV1}. It features 6,637 temporal annotations for events such as goals, cards, and substitutions, facilitating research in action spotting and event detection~\cite{SoccerNetV1}. SoccerNet-v2 expanded this dataset with approximately 300,000 manual annotations across 500 untrimmed broadcasts, introducing tasks like camera shot segmentation and replay grounding~\cite{SoccerNetV2}. The SoccerDB dataset~\cite{SoccerDB2024} offers 171,191 video segments from 346 high-quality matches, supporting object detection, action recognition, temporal action localization, and highlight detection tasks.Additionally, the SoccerReplay-1988 dataset~\cite{matchvision2024} includes 1,988 matches with event labels and textual commentaries, serving as a rich resource for multimodal research. Specialized datasets like SoccerNet-XFoul, annotated by over 70 referees with 22,000 video-question-answer triplets~\cite{SoccerNetXVARS}, and VARS, featuring multi-view clips with detailed foul descriptions~\cite{SoccerNetVARS}, address the complexities of refereeing decisions. These datasets form the backbone for training and evaluating AI systems in soccer video analysis.

\subsection{Multimodal Soccer Video Understanding}
Recent progress in multimodal models has significantly enhanced soccer video understanding~\cite{Survey_soccer_2024, sportu2024, matchvision2024}. Vision-Language Models (VLMs) and Multimodal Large Language Models (MLLMs) now excel in tasks such as classification, segmentation, image-text retrieval, dense video captioning, temporal alignment, and audio description~\cite{VideoUnderstanding_llm, Evolution_of_LML,qwen2-vl}. Notably in soccer, the MatchVision model~\cite{matchvision2024}, leveraging spatiotemporal information, supports event classification, commentary generation, and multi-view foul recognition within a unified framework. Similarly, the SPORTU benchmark~\cite{sportu2024} evaluates MLLMs across multi-level sports reasoning tasks, including rule comprehension and strategic analysis. The X-VARS model~\cite{SoccerNetXVARS} addresses video-based question answering, action recognition, and conversational interactions tailored for professional soccer refereeing contexts. These models demonstrate the potential of integrating visual, textual, and auditory data for comprehensive soccer video understanding, although challenges such as domain-specific adaptation and explainability remain~\cite{SoccerNetXVARS, sportu2024}.

\subsection{Multimodal Integration Techniques}
Effective multimodal integration techniques are vital for robust soccer video understanding~\cite{Gautam2023OctBridging}. The fusion of visual and textual data enables contextual comprehension, with models embedding visual information from images and videos into language model spaces~\cite{Trends_vision_nlp,trend_multimodal,VideoUnderstanding_llm}. The SoccerReplay-1988 dataset~\cite{matchvision2024}, combined with the visual-language foundation model MatchVision, exemplifies successful spatiotemporal integration, excelling in event classification and commentary generation. Unified frameworks like SCBench and its associated CommentarySet dataset~\cite{sportu2024} assess the performance of video-language models in generating relevant commentary. Moreover, multimodal deep learning approaches utilizing video, static images, audio, and optical flow data have been employed for real-time event detection, enhancing model performance by addressing labeling inconsistencies and image quality issues~\cite{amazon2021multimodal,Xiao2022JulOptical, Shaikh2024Apr, Multimodal_anomaly}. These integration techniques are essential for developing AI systems capable of delivering holistic insights in sports video analysis~\cite{spoprts_Breakthroughs}.

\subsection{Datasets for Fine-Tuning Multimodal Models}
Instruction datasets play a crucial role in fine-tuning multimodal language models for soccer video analysis~\cite{Survey_soccer_2024, SoccerNetXVARS}. TaskGalaxy, introduced in 2025, comprises over 19,000 hierarchical task types~\cite{TaskGalaxy}, enabling models to understand complex multimodal instructions. The SPORTU benchmark further supports this by offering SPORTU-text, featuring 900 multiple-choice questions with human-annotated explanations, and SPORTU-video, containing 1,701 slow-motion clips with 12,048 QA pairs~\cite{sportu2024} . These resources evaluate the reasoning abilities of multimodal models by integrating textual and visual information~\cite{ability_Multimodal,MultimodalIntelligence}. Fine-tuning typically involves supervised learning, where models learn task-specific patterns and instructions from these comprehensive datasets, significantly improving performance in sports video understanding tasks~\cite{Madan2024May,VideoUnderstanding_llm,Sportify,Maaz2024Jun}.

\subsection{Multimodal AI in Soccer Analysis}
The applications of multimodal AI in soccer analysis are diverse and transformative~\cite{AiCommentator,Survey_soccer_2024}. Real-time event detection benefits from deep learning models that combine video, audio, static images, and optical flow data~\cite{Yu01122020,Murthy2020May,Afyouni2022Mar}. Commentary generation systems, evaluated by benchmarks like SCBench and CommentarySet dataset, provide contextually relevant narratives~\cite{SCBench}. Tactical analysis systems utilize player tracking and state reconstruction to understand formations and strategies~\cite{TacticAI,systematic_tracking_review}, while sports health monitoring assesses athlete performance and workload through video analysis~\cite{Ferraz2023Nov,Gomez-Carmona2020Aug}. Intelligent refereeing systems, such as those developed using the SoccerNet-XFoul dataset~\cite{SoccerNetXVARS}, integrate multimodal data to support complex decision-making processes. Generative AI is also revolutionizing sports content creation and post-production, allowing broadcasters to automate routine tasks, generate match highlights rapidly, and engage fans through social media~\cite{SoccerStorytelling,Sattar,ibm2024fanengagement}. These applications underscore the potential of multimodal AI to enhance the depth, accuracy, and interactivity of soccer video analysis.

\subsection{Challenges and Future Directions}
Despite significant advancements, several challenges persist in multimodal AI for soccer video understanding~\cite{SportsIntelligence,matchvision2024}. Adapting vision-language models to specific professional domains like sports refereeing remains complex, requiring explainability and domain-specific knowledge~\cite{SoccerNetVARS,matchvision2024,sportu2024}. Real-time analysis demands robust models capable of handling multimodal data with minimal latency~\cite{Hossain2023Oct}. The integration of generative AI and multimodal learning presents opportunities for transforming content creation and fan engagement, but requires careful handling of data quality and contextual accuracy~\cite{Cao2023Mar,Bengesi2024May}. Future research should focus on developing explainable models tailored to professional sports contexts, improving real-time processing capabilities, and enhancing domain-specific adaptations~\cite{Survey_soccer_2024,Miller2024Dec}. The continuous evolution of large-scale, annotated datasets and sophisticated multimodal frameworks will be pivotal in addressing these challenges and advancing soccer video understanding~\cite{SoccerDB2024, SoccerNetV2,matchvision2024}.

\section{Dataset}
\label{sec:dataset}

\begin{figure}[!b]
  \centering
  \includegraphics[width=\linewidth]{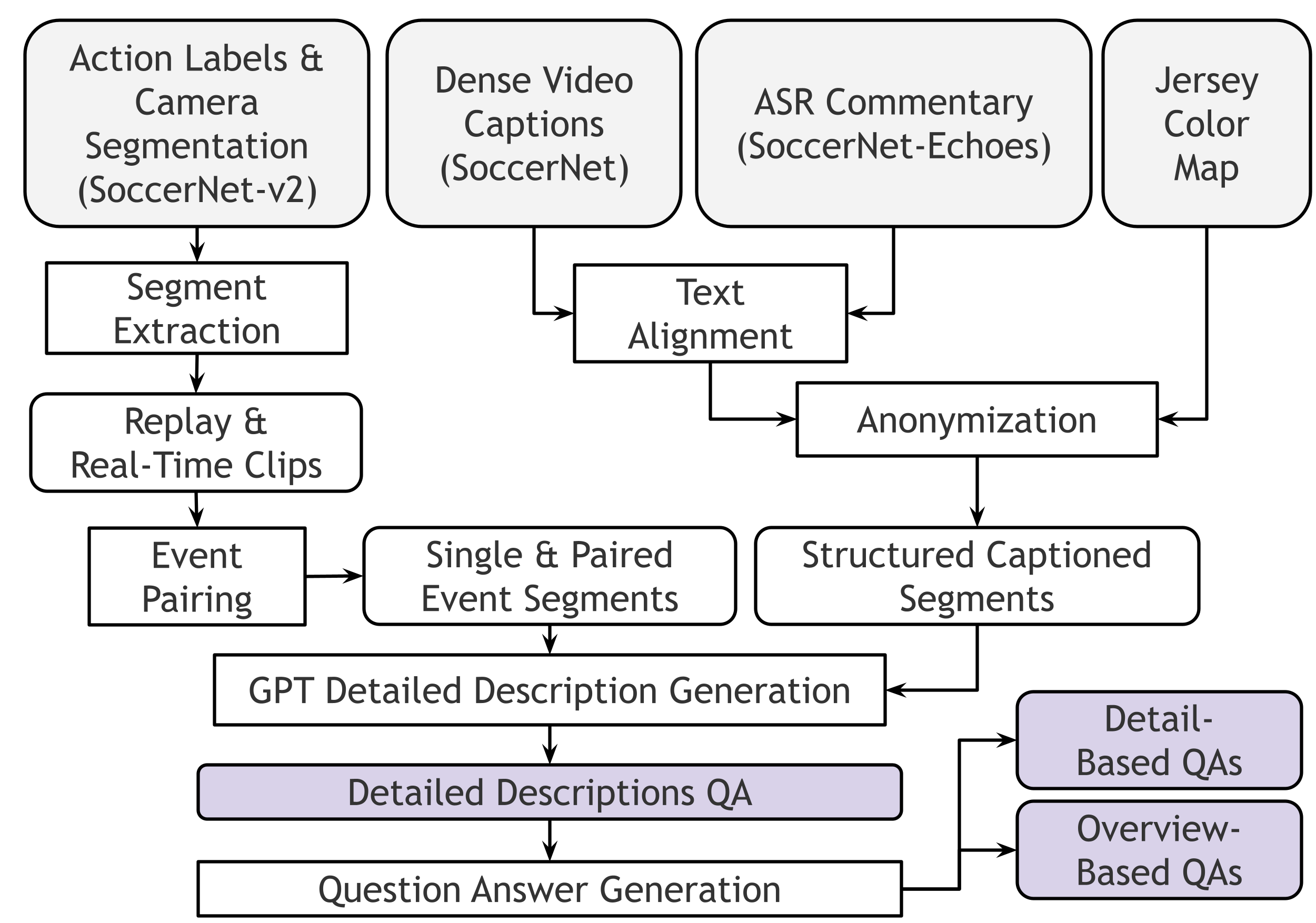}
 \caption{Pipeline that processes different dataset modalities to generate question-answer (QA) pairs for the SoccerChat instruction dataset.}
  \label{fig:SoccerChat-dataset}
\end{figure}

This section outlines the comprehensive methodology employed for creating, refining, and analyzing a soccer video chat instruction dataset derived from multiple existing soccer datasets. Each step has been explained to ensures reproducibility, accuracy, and scientific rigor, enabling an in-depth review of the method used here for the multimodal soccer video understanding and have been summarized in Figure \ref{fig:SoccerChat-dataset}.

\subsection{Dataset Sources}
The datasets utilized in this work include:

\begin{enumerate}
    \item \textbf{SoccerNet-v2}~\cite{SoccerNetV2}: A large-scale dataset designed for analyzing broadcast soccer videos. It comprises 500 full-length matches from major European leagues, totaling approximately 764 hours of footage. Our focus is on two key aspects:
    \begin{itemize}
        \item \textbf{Event Labels}: Annotated action labels identifying specific soccer events (e.g., goals, fouls, substitutions) along with timestamps. The dataset includes around 300,000 manually annotated events covering 17 action classes, essential for associating key moments with corresponding video segments.
        \item \textbf{Camera Shot Segmentation}: Annotations specifying camera types and shot transitions ("replay" or "real-time"), along with their start and end times. These annotations facilitate identifying transitions between replays and real-time gameplay within broadcast footage.
    \end{itemize}
    
    \item \textbf{SoccerNet Dense Video Captioning}~\cite{SoccerNetV2DenseVideo}: Provides timestamped textual captions describing key soccer match events. These captions, sourced from online game portals, serve as a foundation for generating video segment descriptions and question-answer pairs. The dataset includes references to players and teams by name.
    
    \item \textbf{SoccerNetEchoes}~\cite{SoccerNetEchoes}: Contains automated speech recognition (ASR) transcripts from human commentary during soccer matches. While ASR output may introduce noise and references to elements not visible in the video or unrelated to the current match, it enhances contextual understanding of events.
    
    \item \textbf{Jersey Color Information}: Annotated jersey colors of home and away teams across all games are used to generalize player and team identifiers. Since captions and commentary contain real player and team names, jersey color data anonymizes these references. This ensures that models focus on event dynamics rather than memorizing specific names, facilitating more generalizable representations.
\end{enumerate}

\subsection{Dataset Construction}

The construction of the soccer video instruction dataset involved a multi-step process, leveraging the aforementioned datasets to segment the video and align information in different datasets to the segments effectively.

\subsubsection{Soccer Clip Extraction}
To extract relevant video clips, the SoccerNet datasets were utilized. 
The segmentation process involved identifying intervals labeled as either "replay" or "real-time" from Camera Shot Segmentation annotations, ensuring that each clip included relevant actions without any camera transitions.
 Action labels from SoccerNet-v2 were then paired with real-time segments, ensuring consistency in camera type within a $\pm5$-second window centered on action events. 
 Replay segments were subsequently paired with corresponding real-time segments by identifying transition points. These paired segments were refined to ensure that the final clips were no longer than 10 seconds, with truncation applied as necessary.
This resulted in a dataset comprising 90,834 event clips. Metadata for these clips, including their labels, timestamps, camera annotations, and replay type was also extracted. \textit{ffmpeg} was used to crop the game videos into smaller clips.

\subsubsection{Identifying Consecutive Events}
To produce short clips containing two events, we first identified consecutive events with temporal proximity. Specifically, we selected events that occurred between 1 and 7 seconds after the preceding event. Each clip segment was then annotated as "start," "end," or "unrelated" based on its role within the pair. Next, we validated the segments to ensure they were not simply replays of the same event. Valid event pairs were further refined to ensure their boundaries fit within a maximum duration of 10 seconds, and segments longer than 8 seconds were additionally flagged. This process yielded 12,827 valid event pairs, each with a calculated time difference.

\subsection{Caption and Commentary Integration}

To enhance the audio-visual modality with textual context, captions from SoccerNet Dense Video Captioning and commentary from SoccerNetEchoes were integrated. Specific player and team names were replaced with jersey color references (e.g., "red-jerseyed team") to maintain anonymity. The captions were aligned with video segments by matching them within a window that begins 3 seconds after an event and ends 10 seconds after the event. This window was determined empirically, reflecting the time required for humans to process game events before producing captions or commentary.

ASR commentaries were filtered to exclude replay segments as well as redundant words, reducing the likelihood of hallucinations by language models in the ASR. The filtering process resulted in 10,615 single-event clips and 2,982 paired-event clips with associated captions and commentary. This integration provided a rich textual layer to accompany the video data.

\subsection{Data Fusion and Question-Answer Generation}
\label{sec:Data_Fusion}
\subsubsection{Structured Descriptions with GPT}
Multimodal data fusion was employed to generate structured descriptions and question-answer (QA) pairs for each soccer video clip. Event information, jersey colors, captions, and commentary were integrated into templated descriptions, ensuring contextual consistency.

\textbf{Long Description Generation}\\
To anonymize team and player identifiers, jersey colors replaced original names in captions, while commentary retained its original identifiers. Prompt engineering guided the language model to exclude name-based references in generated descriptions, ensuring uniformity. Descriptions strictly adhered to visual content, avoiding direct mention of captions or commentary as sources.

Detailed descriptions, typically 250–300 words, were generated using GPT-3.5 Turbo , focusing solely on observed events. These descriptions captured essential contextual details and served as foundational content for QA generation. Each was paired with a long-form question encouraging inference from visual data rather than textual annotations. Structural variations between single-event and paired-event clips were addressed through tailored prompts and scripts, ensuring consistency in outputs.

\textbf{Overview-Based and Detail-Based QAs}\\
\textit{Overview-based QA} provided a high-level synthesis of visible events, emphasizing the game's broader narrative rather than granular details. Questions focused on overall flow, strategic developments, or key moments within a clip, ensuring responses reflected general context.

Conversely, \textit{detail-based QA} was applied to paired events, prioritizing precise, fact-based responses. These QAs leveraged temporally related segments to formulate questions about events occurring before or after a given moment. Unlike the broader overview-based approach, detail-based QAs focused on specific details while maintaining contextual alignment with annotated data.

Long descriptions were systematically leveraged as source material, with carefully designed prompts ensuring a structured balance between breadth and depth. Responses were constrained to short, well-formed sentences for clarity and informativeness. This dual approach enriched the dataset by capturing both high-level understanding and granular insights, enhancing annotation comprehensiveness.

\section{Methodology}
\label{sec:methodology}

In this section, we present the architecture and implementation details of SoccerChat, a model fine-tuned from Qwen2-VL to enhance comprehension of soccer-related dialogues using a novel video instructional dataset.

\subsection{SoccerChat Architecture}

\begin{figure}[!b]
  \centering
  \includegraphics[width=\linewidth]{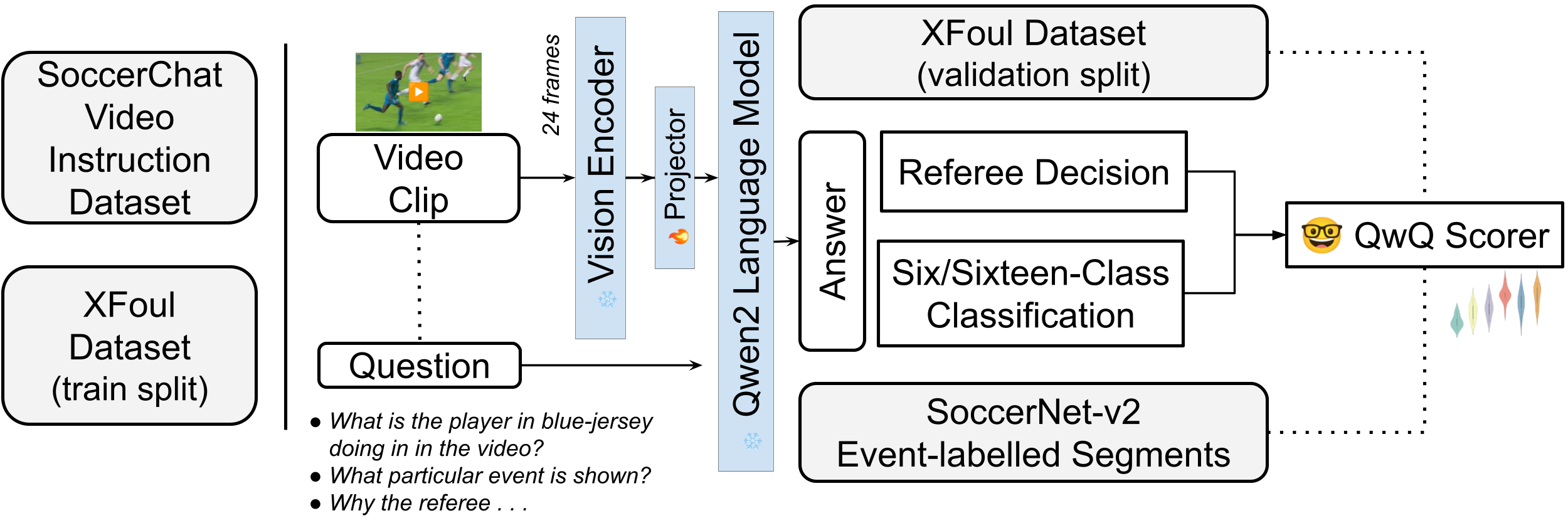}
  \caption{SoccerChat Model based on Qwen2-VL~\cite{qwen2-vl}, illustrating the integration of visual and textual data for enhanced soccer video comprehension.}

  \label{fig:SoccerChat-model}
\end{figure}

SoccerChat builds upon the robust foundation of Qwen2-VL, incorporating specific adaptations to effectively process and understand soccer video content.

\subsubsection{Base Model: Qwen2-VL-7B-Instruct}

Qwen2-VL-7B-Instruct is an advanced vision-language model integrating a Vision Transformer (ViT) with approximately 675 million parameters and the Qwen2 language model, as shown in Figure \ref{fig:SoccerChat-model}. This architecture enables seamless handling of both image and video inputs, facilitating comprehensive multimodal understanding. 

The model incorporates Naive Dynamic Resolution, which dynamically maps images of varying resolutions into a corresponding number of visual tokens, ensuring consistency between input data and inherent visual information \cite{qwen2-vl}. Additionally, it employs Multimodal Rotary Position Embedding (M-RoPE) to decompose positional embeddings into temporal and spatial components, effectively capturing and integrating 1D textual, 2D visual, and 3D video positional information, thereby enhancing its multimodal processing capabilities \cite{qwen2-vl}.

\subsubsection{Fine-Tuning for Soccer Video Processing}

To tailor Qwen2-VL for soccer-specific applications, SoccerChat incorporates the following modifications:

\begin{itemize}
    \item \textbf{Dynamic Resolution Handling for Sports Videos:} Soccer videos in SoccerNet has varying aspect ratios and resolutions. SoccerChat processes images in video frames by resizing them to maintain their original aspect ratio within a maximum pixel dimension of 28×28 per visual patch. This approach ensures that each image is represented by up to 128 visual tokens, preserving the integrity of visual information without distortion. For instance, a video with a 16:9 aspect ratio can be resized to fit within the 28×28 patch grid, resulting in 112 patches, thereby staying within the 128-token limit.
    \item \textbf{Frame Sampling Strategy:} Efficient temporal representation is crucial for understanding dynamic sports content. SoccerChat always extracts a total of 24 frames for each video segment, employing a frame sampling rate of 2.4 frames per second (FPS) for 10-second segment. For example for a segment of 16:9 aspect ratio, this results in a comprehensive token sequence of 24 frames × 112 tokens per frame = 2,688 tokens per segment , capturing essential temporal dynamics while maintaining computational efficiency.
\end{itemize}

While audio can provide valuable contextual information in sports videos, the current implementation of SoccerChat does not directly incorporate audio data, as Qwen2-VL does not support audio processing.

\subsection{Model Configurations and Benchmarking}

The SoccerChat training dataset was constructed by  combining
Long description, Overview-Based and Detail-Based QAs.
To assess the effectiveness of our dataset in training a multimodal soccer video understanding model, we developed and evaluated six model variants. These models were designed to test different training data compositions and configurations to identify the most effective approach for soccer video analysis.
\textbf{SoccerChat} is a Qwen2-VL-7B-Instruct model fine-tuned on the SoccerChat dataset, utilizing structured descriptions and both overview- and detail-based QAs. \textbf{SC+XF} extends SoccerChat by incorporating the XFoul dataset during training to enhance understanding of fouls and referee decisions. \textbf{SC-FT-XF} follows a sequential fine-tuning approach, first on SoccerChat and then on XFoul, evaluating whether this improves performance over direct training. \textbf{Q2VL} serves as a general-purpose baseline, representing the publicly available Qwen2-VL-7B-Instruct model. \textbf{Q2VL-XF} is a public checkpoint of the X-VARS model trained on XFoul. Finally, \textbf{X-VARS} is fine-tuned solely on XFoul, establishing a baseline for referee decision tasks.

\subsection{Evaluation Tasks}

The trained models were evaluated on multiple tasks to measure their effectiveness in different aspects of soccer video understanding. These tasks included question-answer validation using the XFoul dataset and classification tasks using SoccerNet-v2.

\subsubsection{XFoul Question-Answer Validation}

For question-answer validation, we utilized the XFoul validation set as the ground truth. The answers provided by each model were compared against natural sentence responses in the XFoul dataset. The objective was to assess how well the models generate informative and contextually accurate responses.

\subsubsection{Action Classification Tasks}

To evaluate the models’ ability to classify soccer actions, we designed two classification tasks using video segments from SoccerNet-v2. The first task, \textbf{Six-Class Classification}, involved categorizing video clips into six predefined labels: Ball out of play, Foul, Goal, Shots off target, Shots on target, and Throw-in. Models were required to predict the appropriate class and justify their selection, focusing on key events from single-event segments. The second task, \textbf{Sixteen-Class Classification}, extended the complexity by incorporating all labels from SoccerNet-v2, with up to 100 videos per class. This setup included additional labels appearing in segments with paired events, providing a more comprehensive evaluation of the models' classification capabilities.

\subsection{Evaluation Method}
\label{sec:evaluation_method}

To quantitatively measure performance, we employed the QwQ Scorer model, which assigned scores between 0 and 10 based on alignment with ground truth annotations. 0 represents no alignment or correctness, 10 signifies perfect alignment and justification, and scores in between indicate varying levels of partial correctness and alignment.
For the classification tasks, models were prompted to predict the appropriate class label and justify their choice. The generated responses were scored based on their correctness and explanatory quality. For the XFoul validation task, the models’ answers were compared against the XFoul ground truth responses, with scoring based on similarity and contextual alignment.

To visually interpret the performance across different models, results were analyzed and presented using grouped violin plots. These plots illustrated the score distributions, enabling a comparative analysis of the model variants and their effectiveness in various evaluation tasks.

\section{Experiments}
\label{sec:experiments}

We conducted extensive experiments to evaluate the effectiveness of our \textit{SoccerChat} dataset and the proposed models on two primary tasks: \textit{XFoul question-answer validation} (referee decision-making) and \textit{action classification} using SoccerNet-v2. The evaluation relied on the \textit{QwQ Scorer} model, which assigns scores between 0 and 10 based on alignment with ground truth annotations. Performance comparisons were visualized using grouped violin plots to highlight score distributions across different model configurations.

\subsection{XFoul Question-Answer Validation (Referee Decision Task)}

\begin{figure}[!b]
  \centering
\includegraphics[width=0.75\linewidth]{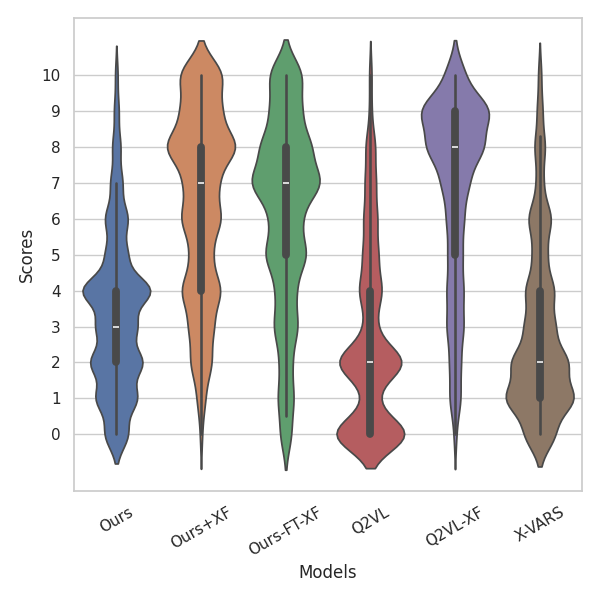} 
  \caption{Score distribution of models for referee decision tasks shown using violin plots with quartiles and median indicators}
\label{fig:xfoul_validation}
\end{figure}

We evaluated multiple models on their ability to answer soccer foul-related questions using the XFoul validation dataset. The score distribution of different model variants for both tasks is shown in Figure \ref{fig:xfoul_validation}.
The Q2VL-XF model achieved the highest average score of 6.81, demonstrating strong alignment with referee decisions. The SC+XF model followed closely with a score of 6.46, indicating that integrating SoccerChat and XFoul datasets enhances referee decision performance. The SC-FT-XF model scored slightly lower at 6.14, suggesting that fine-tuning on XFoul after pretraining on SoccerChat is less effective than training on both datasets simultaneously. The SoccerChat model, which lacked XFoul training, scored significantly lower at 3.37, highlighting its limited ability in referee decision despite broader soccer understanding. Surprisingly, X-VARS, despite being exclusively trained on XFoul, achieved only 2.91, comparable to Q2VL (2.58), which had no soccer-specific training. This suggests that X-VARS may lack robust general pretraining, limiting its effectiveness despite exposure to XFoul data. Standard deviation across XFoul-trained models remained around 2.6, indicating some response inconsistency. Overall, models trained on XFoul significantly outperformed those without, emphasizing the dataset’s importance in referee decision tasks.

\subsection{Action Classification Tasks}
\begin{figure}[ht]
  \centering
\includegraphics[width=\linewidth]{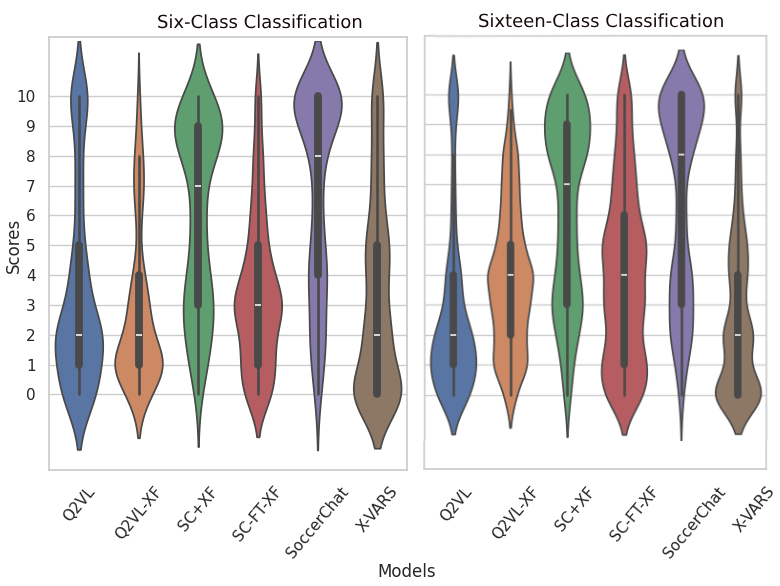} 
\caption{Score distribution of models for six-class (left) and sixteen-class (right) classification tasks.}
\label{fig:SoccerChat_classification_scored}
\end{figure}

We evaluated multiple models on two classification tasks—six-class and sixteen-class event classification—to assess their performance in detecting key soccer events. The score distribution of different model variants for both tasks is shown in Figure \ref{fig:SoccerChat_classification_scored}.

\subsubsection{Six-Class Classification}
\label{sec:Six-Class-Classification}

\begin{figure}[htbp]
  \centering
\includegraphics[width=\linewidth]{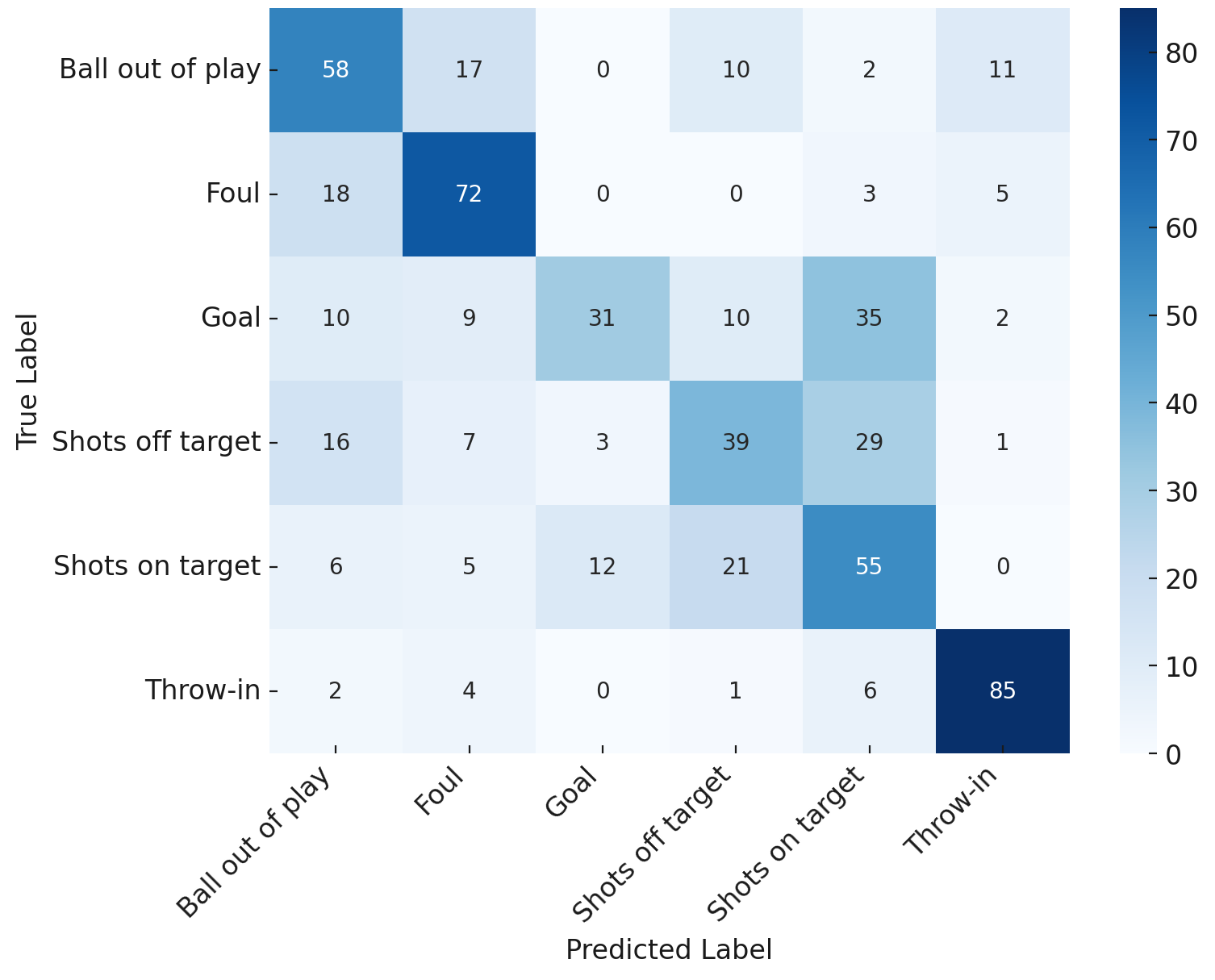} 
\caption{Confusion matrix for the six-class classification task.}
\label{fig:SoccerChat_6cl_confusion}
\end{figure}

\begin{table}[htbp]
  \caption{Summary of the performance of different model variants in the six-class classification task.}
  \scriptsize
  \centering
  \footnotesize
  \renewcommand{\arraystretch}{1.2} %
  \begin{tabular}{lccccccc}
      \toprule
      Model &
      \rotatebox{90}{Precision (wt)} & 
      \rotatebox{90}{Recall (wt)} & 
      \rotatebox{90}{F1 Score (wt)} & 
      \rotatebox{90}{Cohen Kappa} & 
      \rotatebox{90}{MCC} & 
      \rotatebox{90}{Hamming Loss} \\
      \midrule
      \textbf{SoccerChat} & 0.59 & 0.57 & 0.57 & 0.48 & 0.49 & 0.43 \\
      SC+XF & 0.56 & 0.49 & 0.49 & 0.39 & 0.41 & 0.51 \\
      Q2VL & 0.47 & 0.20 & 0.17 & 0.10 & 0.13 & 0.80 \\
      X-VARS & 0.56 & 0.21 & 0.19 & 0.11 & 0.14 & 0.79 \\
      SC-FT-XF & 0.39 & 0.17 & 0.12 & 0.02 & 0.03 & 0.83 \\
      Q2VL-XF & 0.03 & 0.17 & 0.05 & 0.00 & 0.01 & 0.83 \\
      \bottomrule
  \end{tabular}
  \label{tab:performance_metrics_6}
\end{table}

The evaluation of six model variants for soccer action classification, as shown in Table \ref{tab:performance_metrics_6}, revealed significant performance differences. 

As shown in Figure \ref{fig:SoccerChat_classification_scored} (left), SoccerChat emerged as the strongest performer, achieving the highest mean score (6.80) with a 75th percentile score of 10, indicating strong alignment with ground truth annotations. SC+XF, trained on both SoccerChat and XFoul datasets, also performed well with a mean score of 5.97, suggesting that foul-related data integration did not significantly compromise general event classification. However, SC-FT-XF underperformed with a mean score of 3.40, implying that fine-tuning on foul-related data diminished its general classification ability. General-purpose models, Q2VL (3.44) and X-VARS (3.26), struggled, reinforcing that models not specifically trained on soccer-related datasets lack precise event classification capabilities. The weakest performer, Q2VL-XF (2.72), demonstrated that training exclusively on XFoul fails to generalize to broader soccer actions. These results highlight the importance of balanced dataset exposure for optimizing classification accuracy. The confusion matrix for the validation set in Figure \ref{fig:SoccerChat_6cl_confusion} shows that the SoccerChat model correctly classifies most football events, particularly corners, penalties, and yellow cards. However, it struggles with similar categories, such as direct vs. indirect free-kicks and shots on vs. off target.

\subsubsection{Sixteen-Class Classification}
The sixteen-class classification task further evaluated six models using the QwQ Scorer. As shown in Figure \ref{fig:SoccerChat_classification_scored} (right), SoccerChat achieved the highest mean score of 6.42, reinforcing its effectiveness in soccer action classification. SC+XF followed with a score of 6.15, slightly trailing SoccerChat, indicating that integrating XFoul data introduced minor inconsistencies. SC-FT-XF (4.00) performed notably worse, suggesting that fine-tuning on foul-related data negatively impacted overall soccer event understanding. General-purpose models, Q2VL (2.72) and X-VARS (2.70), recorded the lowest scores, confirming their struggle in soccer-specific contexts. Q2VL-XF (3.88), trained on XFoul, slightly improved upon Q2VL but remained significantly behind soccer-specialized models. The overall trend demonstrated that models extensively trained on the SoccerChat dataset (SoccerChat, SC+XF) outperformed those focused solely on fouls (X-VARS, Q2VL-XF) or general-purpose vision-language models (Q2VL).

\section{Discussion}
\label{sec:discussion}

The evaluation results across different tasks consistently highlight the critical role of training methodology, dataset specialization, and the balance between general pretraining and fine-tuning. The Q2VL-XF model achieved the highest performance in the XFoul question-answer validation task due to its direct training on the XFoul dataset, allowing it to capture intricate referee decision-making patterns. Similarly, SC+XF performed well, demonstrating that joint training on SoccerChat and 	XFoul enables models to understand both general soccer discussions and referee decisions effectively. In contrast, 	SC-FT-XF, which was sequentially fine-tuned on XFoul after SoccerChat, performed worse, suggesting that training datasets together from the outset is a more effective strategy than fine-tuning in stages. 

The generally poor performance of SoccerChat and X-VARS in the XFoul validation task suggests that while SoccerChat is strong in soccer-related discussions, it lacks referee decision-specific knowledge. Similarly, despite being trained exclusively on XFoul, X-VARS underperformed due to its lack of a strong pretraining foundation. These findings emphasize the importance of robust pretraining before fine-tuning on specialized tasks. Additionally, general-purpose vision-language models like Q2VL struggled the most, reinforcing the need for domain-specific training to achieve strong soccer analysis performance.

The classification tasks further reinforced these insights. SoccerChat excelled in soccer event classification due to its structured dataset, while SC+XF benefited from incorporating foul-specific knowledge without losing general classification ability. However, SC-FT-XF underperformed, likely due to overfitting on fouls, reducing its ability to generalize to other soccer actions. Models trained exclusively on foul datasets, such as X-VARS and Q2VL-XF, struggled with broader soccer classification, indicating that recognizing fouls alone does not translate into a comprehensive understanding of soccer dynamics.
The sixteen-class classification task further emphasized that fine-tuning on XFoul after SoccerChat training could lead to catastrophic forgetting, where the model becomes overly specialized in fouls at the cost of broader soccer comprehension. Interestingly, SC+XF slightly underperformed compared to SoccerChat in this task, suggesting that the inclusion of XFoul data might introduce conflicting decision patterns, making the model more cautious in general classification.

Overall, these experiments confirm that optimal performance in soccer-related AI tasks requires both strong general pretraining and carefully structured domain-specific training. Joint training on multiple datasets proves to be more effective than sequential fine-tuning, ensuring a balanced and contextually aware understanding of both general soccer events and referee decision. These findings highlight the importance of dataset selection and training strategy when designing AI models for specialized sports analytics tasks.

\section{Conclusion}
\label{sec:conclusion}
In this paper, we presented SoccerChat, a multimodal AI framework designed to enhance soccer game understanding through the integration of visual, auditory, and textual data. By enriching the SoccerNet dataset with jersey color annotations and ASR transcripts, we created a robust instruction dataset to train and evaluate SoccerChat on key soccer analytics tasks. Experimental results show that joint training on multimodal data significantly improves event understanding and referee decision making, outperforming general-purpose models and fine-tuned baselines. However, sequential fine-tuning on specialized datasets introduces trade-offs in generalizability, underscoring the need for carefully structured dataset integration. Our work demonstrates the potential of multimodal AI in sports analytics, setting the stage for future research in real-time soccer analysis, interactive sports AI, and explainable decision-making systems.

{\small
\bibliographystyle{ieeenat_fullname}
\bibliography{main}}
\clearpage
\clearpage
\setcounter{page}{1}
\maketitlesupplementary
\section{SoccerChat Dataset Generation}

This section presents the dataset generation templates, which facilitate long description generation and short question-answering based on soccer video clips. 

Figure~\ref{list:Data_Fusion_example_e1} illustrates the instruction template used for generating long descriptions of video clips containing a single event. The instructions define the LLM's role in generating detailed and well-structured descriptions based on fragmented information about a soccer clip. The template ensures that the response is purely visual-based and does not rely on commentary or captions explicitly. The input format for this generation task for GPT-3.5 Turbo is presented in Figure~\ref{list:Data_Fusion_example_e1_sample}, which structures the event, jersey colors, supporting captions, and commentary into a structured input format.

For generating detailed, event-based QAs, Figure~\ref{list:Detail-Based-QAs-template} provides a structured guideline. The system first plays the role of a human inquiring about specific soccer events and then acts as an AI assistant providing detailed responses. The generated questions focus on extracting event-specific details directly from the provided captions while ensuring that answers remain descriptive and insightful.

Table~\ref{tab:soccerchat_examples} provides randomly selected examples of question-answer pairs from the SoccerChat dataset. Each example corresponds to a specific soccer video clip, such as a corner kick or a throw-in leading to a shot off target. The questions are designed to explore various aspects of the events, including player actions, defensive tactics, and the overall strategic context of the game. The answers provide comprehensive insights into the visual content of the clips, simulating a detailed soccer analysis.

These structured templates and examples contribute to enhancing automatic video understanding by generating high-quality descriptive text and detailed QAs for soccer game clips.

\section{QwQ Scorer}

To ensure a structured and consistent evaluation of model-generated classifications, we employed the QwQ Scorer as described in Section \ref{sec:evaluation_method}, which assigns a score between 0 and 10 based on alignment with ground truth annotations. The scorer considers both the correctness of the classification and the quality of the justification. Listing \ref{list:qwq_scorer_template} provides the structured template used for scoring classification tasks particularly for six-class classification as in \ref{sec:Six-Class-Classification}. This template defines the evaluation criteria, input format, and the expected structured output in the form of a Python dictionary. The scorer ensures that no two models receive the same score, enhancing the differentiation in performance comparison. Additionally, it enforces alignment with six predefined class labels—‘Ball out of play’, ‘Foul’, ‘Goal’, ‘Shots off target’, ‘Shots on target’, and ‘Throw-in’—with an explicit ‘Wrong Prediction’ category for misclassified outputs.

\begin{figure}[htbp]
\lstset{breaklines=true}
    \begin{lstlisting}[basicstyle=\ttfamily\footnotesize, caption={Input template for the long description generation particularly for clips with single event in Listing \ref{list:Data_Fusion_example_e1}. The angle brackets hold variables received from the real dataset with round brackets showing the data source.}, label={list:Data_Fusion_example_e1_sample}]
Video Clip:  
shows only <(event)Goal> event by 
<(home_jersey_color)red>-jerseyed team in the match between teams in 
<(home_jersey_color)red> vs 
<(away_jersey_color)blue> jerseys.  
-----  
Possible Supporting Caption:  
<(caption)red-jerseyed team scores a brilliant goal!>  
-----  
Possible Supporting Commentary:  
<(ASR)What a stunning finish! The home team takes the lead!>  
    \end{lstlisting}
\end{figure}

\begin{figure*}[htbp]
\lstset{breaklines=true}
    \begin{lstlisting}[basicstyle=\ttfamily\footnotesize, caption={Instructions for long description generation particularly for clips with single event as described in Section \ref{sec:Data_Fusion}.}, label={list:Data_Fusion_example_e1}]
{
    "role": "system",
    "content": 
        "You will play two roles: a human asking questions related to describing a short soccer video clip and an intelligent chatbot designed for video description, storytelling and captioning. Your task is to generate a detailed and descriptive paragraph based on the provided fragmented information about a short video clip. "
        "##TASK:"
        "Users will provide event description, supporting caption and commentary of a clip, and you will generate ONE conversation-like question and answer related to describing the video and the game event in detail. The question should ask to describe the video content in detail. The answer should be a paraphrased and well-structured paragraph based on the provided description, with a minimum of 250 words and a maximum of 300 words. "
        "##INSTRUCTIONS:"
        "- The question must be like a human conversation and focused on describing the video and event in detail. "
        "- Reject the information in supporting commentary and caption if not relevant and logical to the event visible in the clip. "
        "- The answer must be a paraphrased version of the provided information, very detailed and descriptive, and within the specified word count. "
        "- Act as if you are really seeing the visual content live and have no access to the commentary and caption. Don't mention about 'commentary' and 'caption' in the answer. "
        "- Only use the supporting commentary and caption to be smart enough to interpret the visual content, faking as though you got the information from the video itself."
        "- Avoid mentioning actual player names and team names from the commentary as it is not visible in video; instead, refer to them by jersey-color if possible, else ignore the information."
        "- Begin answers with creative opening."
},
{
    "role": "user",
    "content":
        f"The fragmented information: {game-details}. Please generate the response in the form of a Python JSON dictionary string with keys 'Q' for question and 'A' for answer. Each corresponding value should be the question and answer text respectively. "
        "For example, your response should look like this: {'Q': 'Your question here...', 'A': 'Your answer here...'}. Emphasize that the answer should focus on describing the video content as detailed as possible."
}
    \end{lstlisting}
\end{figure*}

\begin{figure*}[htbp]
\lstset{breaklines=true}
    \begin{lstlisting}[basicstyle=\ttfamily\footnotesize, caption={Instructions for long detail-based QAs generation particularly as described in Section \ref{sec:Data_Fusion}.}, label={list:Detail-Based-QAs-template}]

    {
        "role": "system",
        "content": "You play two roles: a human asking questions related to a short soccer video clip and an intelligent chatbot designed to help people understand specific events within the clip. "
        "Your task is to focus on soccer video summarization, which will be utilized by users to comprehend key moments in soccer matches through various questions based on the video content. "
        "This summarization will assist in applications like analyzing game highlights, generating summaries for sports content platforms, creating brief overviews for coaching analysis, or providing quick updates for fans. "
        "You will first act as a human inquiring about specific events in a soccer match and then switch roles to an AI assistant providing detailed information based on the video's content."
        "------"
        "##TASK:"
        "You will be given a caption of a specific event from a short soccer video clip. Based on this caption, you will generate a set of conversational-style questions and answers related to the visible events. <event_info> "
        "The questions should be crafted to extract information DIRECTLY from the provided caption, so that it or parts of it can serve as the answers. "
        "Generate THREE different descriptive and conversational style questions and detailed answers based on the given information."
        "------"
        "##INSTRUCTIONS:"
        "- The questions must be conversational and directly related to the events in the soccer video clip. "
        "- The questions should be designed to extract information DIRECTLY from the given caption, so that it or parts of it can serve as the answers. "
        "- The answers must be detailed, descriptive, and should directly reference the information provided. "
        "- The questions can focus on player actions, game strategies, scoring opportunities, defensive tactics, or any key moments in the clip. "
        "------"
        "##SAMPLE QUESTIONS (based on given caption and event type):"
        "- How did the player score the goal in the clip?"
        "- What defensive strategy did the team use to prevent the goal?"
        "- Describe the sequence of passes that led to the goal."
        "- Was there an offside violation in the buildup to the goal?"
        "- How did the goalkeeper react to the shot?"
    },
    {
        "role": "user",
        "content": "The video caption is: <long_description>. "
        "Please generate the response in the form of a Python JSON, where JSON strings start with keys 'Q' for question and 'A' for answer. Each corresponding value should be the question and answer text respectively. "
        "The response should look EXACTLY like this : {'Q1': 'Your first question here...', 'A1': 'Your first answer here...', 'Q2': 'Your second question here...', 'A2': 'Your second answer here...', 'Q3': 'Your third question here...', 'A3': 'Your third answer here...'}. "
        "Emphasize that ALL THREE questions must be designed to extract information DIRECTLY from the given caption, so that it or parts of it can serve as the answers, and provide detailed and descriptive answers."
    }

    \end{lstlisting}
\end{figure*}

\clearpage
\onecolumn
\begin{longtable}[htbp]{|p{5.5cm}|p{9.5cm}|}
    \caption{Random examples with questions and answers from the SoccerChat dataset} \label{tab:soccerchat_examples} \\
    \hline
    \textbf{Question} & \textbf{Answer} \\ 
    \hline
    \endfirsthead
    
    \hline
    \textbf{Question} & \textbf{Answer} \\ 
    \hline
    \endhead

    \hline
    \multicolumn{2}{|c|}{\multirow{2}{*}{Single Event Clip: \textit{40035\_Corner.mp4}}} \\
    \multicolumn{2}{|c|}{} \\ 
    \hline
    Can you describe the goalkeeper's actions during the corner kick? &
    A player from the blue/red stripe-jerseyed team delivers the corner kick, and the goalkeeper of the yellow jerseys team leaps into action, swiftly coming off his line to intercept the corner and prevent any potential threat. \\ 
    \hline
    What key event is captured in the video clip? & 
    The video captures a corner event in a match between teams wearing blue/red stripe jerseys and yellow jerseys. \\ 
    \hline
    How does the goalkeeper's play reflect the intense tactical battle in the match? & 
    The strategic positioning and decisive action of the goalkeeper reflect the intense tactical battle unfolding in the match, as both teams seek to gain the upper hand and secure a positive outcome. \\ 
    \hline
    
    \multicolumn{2}{|c|}{\multirow{2}{*}{Two Event Clip: \textit{113345\_Throw-in--Shotsofftarget.mp4}}} \\
    \multicolumn{2}{|c|}{} \\ 
    \hline
    How did the goalkeeper from the turquoise/teal-jerseyed team respond to the long-range shot attempted by the red/white stripe-jerseyed player? &
    The goalkeeper from the turquoise/teal-jerseyed team adeptly responded to the long-range shot attempted by the red/white stripe-jerseyed player by showcasing remarkable anticipation and agility. This response involved positioning, reflexes, and decision-making to make a comfortable save, demonstrating the goalkeeper's skill in protecting the goal. \\ 
    \hline
    Could you briefly summarize the video content? & 
    The video showcases a strategic throw-in by the red/white stripe-jerseyed team, followed by a skilled long-range shot towards the goal, which was met with an adept save by the goalkeeper from the turquoise/teal-jerseyed team. It captures the intense battle and the skillful gameplay between the two teams. \\ 
    \hline
    What are the main events shown in the video? & 
    The main events in the video include the red/white stripe-jerseyed team taking a throw-in to showcase their strategic gameplay, followed by a player attempting a long-range shot towards the bottom left corner of the goal, which was skillfully saved by the goalkeeper from the turquoise/teal-jerseyed team. \\ 
    \hline
    What does the skillful play by the red/white stripe-jerseyed team and the reaction by the turquoise/teal-jerseyed team's goalkeeper reveal about the battle between the two teams? & 
    The skillful play by the red/white stripe-jerseyed team and the reaction by the turquoise/teal-jerseyed team's goalkeeper reveal the intense battle between the two teams. The red/white stripe-jerseyed team persistently attempts to breach the opposition's defense through skillful passing and opportunistic long shots, while the turquoise/teal-jerseyed team's goalkeeper remains alert and well-positioned to defend their goal. This highlights the competitive nature of the match and the strategic efforts of both teams to gain an advantage. \\ 
    \hline
    What strategic gameplay did the red/white stripe-jerseyed team showcase during the throw-in? & 
    During the throw-in, the red/white stripe-jerseyed team showcased strategic gameplay by making calculated decisions to maintain possession and advance their position on the field. This may have involved specific player movements, strategic passing options, or tactical positioning to create potential scoring opportunities based on the situation at hand. \\ 
    \hline
\end{longtable}
\clearpage
\twocolumn

\begin{figure*}[t]
\lstset{breaklines=true}
    \begin{lstlisting}[basicstyle=\ttfamily\footnotesize, caption={Template for QwQ Scorer used for performance evaluation particularly for classification as described in Section \ref{sec:evaluation_method}. The scorer assigns a score between 0 and 10 based on the alignment of model-generated responses with ground truth annotations. The template provides a structured format for inputting predictions, justifications, and assessing correctness and explanatory quality. This ensures consistent evaluation across classification tasks.}, label={list:qwq_scorer_template}]
    The task is to classify the outputs of different models based on whether they correctly identified the given class. The score reflects the model's ability to correctly classify the Query into the Actual Label provided.

    ### Scoring Criteria:
    - 0: Completely incorrect classification (model's answer does not match the Actual Label at all).
    - 10: Fully correct classification (model's answer exactly matches the Actual Label).

    ### Inputs:
    Query: {query}
    Actual Label: "{label}"
    LLM-Answers: {answers}

    ### Task:
    For each model's answer provided in the LLM-Answers dictionary:
    1. Assess whether the model's output correctly identifies the Actual Label.
    2. Assign a score from 0 to 10 based on the correctness of the classification. If the classification is fully correct, assign a score of 10. If it is completely incorrect, assign a score of 0. Partial correctness or near misses can be scored in between.
    3. Ensure no two models have the same score to clarify the comparison.
    4. predicted_class could be 'Wrong Prediction' if the model's answer is different from the six possible classes

    Output a dictionary enclosed by ``` on both sides with three keys:
    - "scores": A dictionary where each key is the model name from the LLM-Answers dictionary, and the value is the score (range 0-10) assigned to the model's answer.
    - "reason": A dictionary where each key is the model name, and the value is a string explanation for why the score was assigned to the respective model.
    - "predicted_class": A dictionary where each key is the model name, and the value is one best predicted class only out of the six possible classes:  'Ball out of play', 'Foul', 'Goal' , 'Shots off target', 'Shots on target', or 'Throw-in'. Output 'Wrong Prediction' if the  the model's answer does not match any of the possible classes.

    ### Output:
    No coding is required. Directly provide the expected output result as a valid Python JSON dict of dict enclosed in ```.
    \end{lstlisting}
\end{figure*}

\clearpage
\section{More Evaluation Results}
\label{sec:more_evaluation}

\begin{figure}[h]
  \centering
\includegraphics[width=\linewidth]{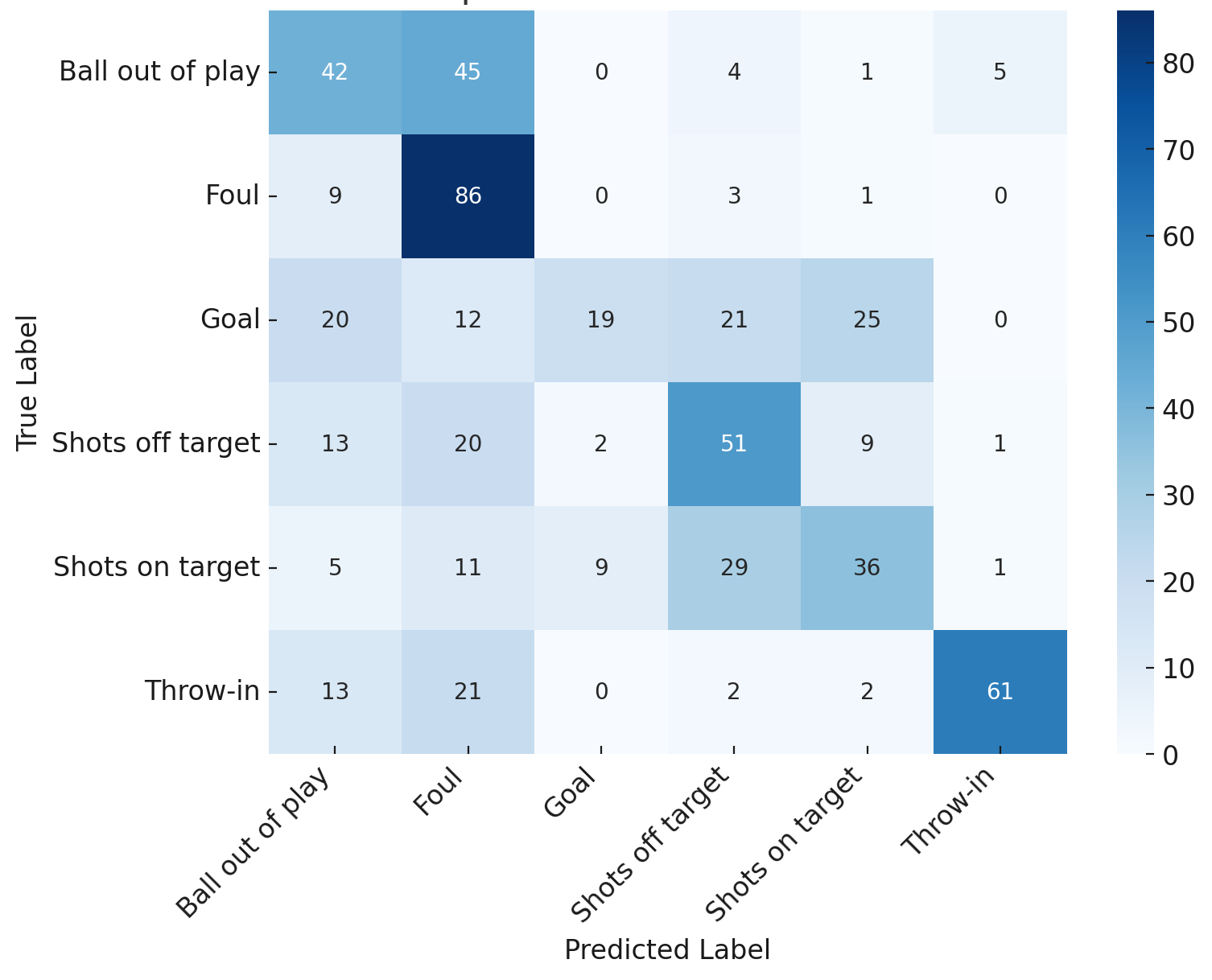} 
  \caption{Confusion matrix for SC+XFoul model for six-class classification problem.}
\label{fig:SoccerChat+XFoul_6cl_confusion}
\end{figure}

\begin{figure}[h]
  \centering
\includegraphics[width=\linewidth]{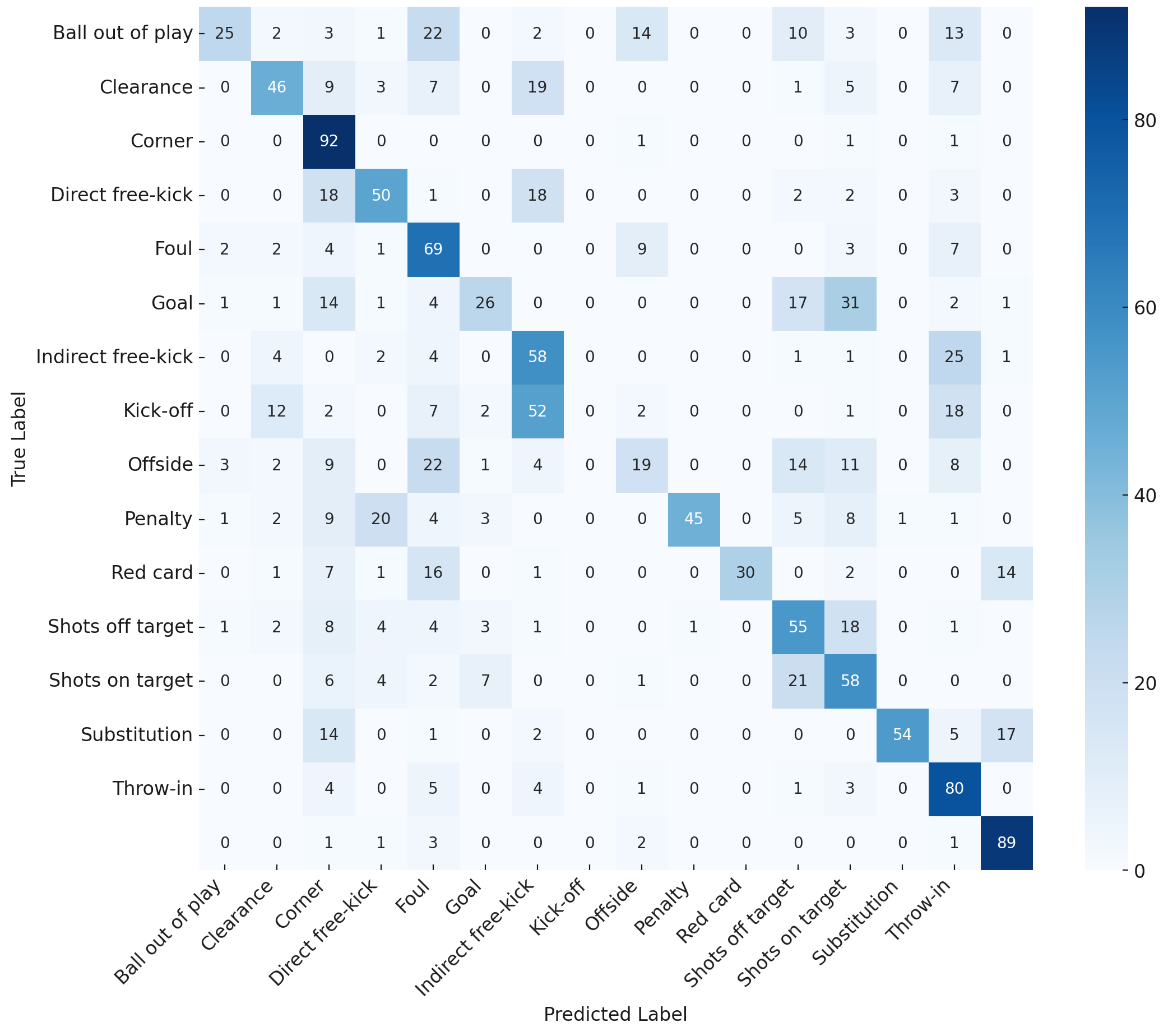} 
  \caption{Confusion matrix for SoccerChat model for sixteen-class classification problem.}
\label{fig:SoccerChat_16cl_confusion}
\end{figure}

\begin{table}[htbp]
  \caption{Classification Report for SoccerChat Model for Sixteen-Class Classification Problem.}
  \scriptsize
    \centering
    \footnotesize
    \begin{tabular}{lcccc}
        \hline
        Class & Precision & Recall & F1 & Support \\
        \hline
        Ball out of play & 0.76 & 0.25 & 0.38 & 100 \\
        Clearance & 0.62 & 0.46 & 0.53 & 99 \\
        Corner & 0.46 & 0.97 & 0.62 & 95 \\
        Direct free-kick & 0.57 & 0.53 & 0.55 & 95 \\
        Foul & 0.40 & 0.69 & 0.51 & 100 \\
        Goal & 0.62 & 0.26 & 0.37 & 100 \\
        Indirect free-kick & 0.36 & 0.60 & 0.45 & 97 \\
        Kick-off & 0.00 & 0.00 & 0.00 & 100 \\
        Offside & 0.39 & 0.19 & 0.26 & 98 \\
        Penalty & 0.98 & 0.45 & 0.62 & 100 \\
        Red card & 1.00 & 0.41 & 0.58 & 74 \\
        Shots off target & 0.43 & 0.55 & 0.48 & 100 \\
        Shots on target & 0.39 & 0.58 & 0.47 & 100 \\
        Substitution & 0.98 & 0.56 & 0.71 & 97 \\
        Throw-in & 0.47 & 0.80 & 0.59 & 100 \\
        Yellow card & 0.73 & 0.91 & 0.81 & 98 \\
        \hline
        Accuracy & &&0.51 & 1553 \\
        Macro avg & 0.54 & 0.48 & 0.47 & 1553 \\
        Weighted avg & 0.57 & 0.51 & 0.49 & 1553 \\
        \hline
    \end{tabular}
    \label{tab:classification_report}
\end{table}

\begin{table}[htbp]
  \caption{Summary of the Sixteen-Class Classification Task.}
  \scriptsize
  \centering
  \footnotesize
  \renewcommand{\arraystretch}{1.2} %
  \begin{tabular}{lccccccc}
      \toprule
      Model &
      \rotatebox{90}{Precision (wt)} & 
      \rotatebox{90}{Recall (wt)} & 
      \rotatebox{90}{F1 Score (wt)} & 
      \rotatebox{90}{Cohen Kappa} & 
      
      \rotatebox{90}{MCC} & 
      \rotatebox{90}{Hamming Loss} \\
      \midrule
      SoccerChat & 0.57 & 0.51 & 0.49 & 0.48 & 0.49 & 0.49 \\
      SC+XF & 0.52 & 0.47 & 0.45 & 0.44 & 0.44 & 0.53 \\
      Q2VL & 0.39 & 0.13 & 0.13 & 0.10 & 0.13 & 0.87 \\
      X-VARS & 0.39 & 0.12 & 0.13 & 0.09 & 0.11 & 0.88 \\
      SC-FT-XF & 0.37 & 0.12 & 0.11 & 0.07 & 0.11 & 0.88 \\
      Q2VL-XF & 0.35 & 0.12 & 0.09 & 0.07 & 0.09 & 0.88 \\
      \bottomrule
  \end{tabular}
  \label{tab:performance_metrics_16}
\end{table}

\begin{table}[htbp]
  \caption{Classification Report for SoccerChat Model for Six-Class Classification Problem.}
  \scriptsize
    \centering
    \footnotesize
    \begin{tabular}{lcccc}
        \hline
        Class & Precision & Recall & F1-score & Support \\
        \hline
        Ball out of play & 0.53 & 0.58 & 0.55 & 100 \\
        Foul & 0.63 & 0.72 & 0.67 & 100 \\
        Goal & 0.67 & 0.31 & 0.42 & 100 \\
        Shots off target & 0.48 & 0.39 & 0.43 & 100 \\
        Shots on target & 0.42 & 0.55 & 0.48 & 100 \\
        Throw-in & 0.82 & 0.85 & 0.83 & 100 \\
        Wrong Prediction & 0.00 & 0.00 & 0.00 & 0 \\
        \hline
        Accuracy & & & 0.57 & 600 \\
        Macro avg & 0.51 & 0.49 & 0.48 & 600 \\
        Weighted avg & 0.59 & 0.57 & 0.57 & 600 \\
        \hline
    \end{tabular}
    \label{tab:classification_report_six_class}
\end{table}

Figure~\ref{fig:SoccerChat+XFoul_6cl_confusion} presents the confusion matrix for the SC+XFoul model on the six-class classification task. The matrix highlights the model’s performance across different event categories, showing patterns in misclassification and the overall distribution of predictions.

Similarly, Figure~\ref{fig:SoccerChat_16cl_confusion} depicts the confusion matrix for the SoccerChat model on the sixteen-class classification problem. This visualization provides insights into how well the model differentiates between various football events, highlighting both strong and weak classification areas. One thing to be noted is that Kick-off event isn't classified at all as examples of these types aren't present in the SoccerChat dataset.

The classification report for the sixteen-class problem, shown in Table~\ref{tab:classification_report}, provides a breakdown of precision, recall, and F1 scores for each event type. Notably, some classes, such as "Yellow card" and "Throw-in," achieve higher recall and F1 scores, whereas others, such as "Kick-off" and "Offside," exhibit lower performance, indicating challenges in their correct classification.

Table~\ref{tab:performance_metrics_16} summarizes the overall performance of different models in the sixteen-class task. The SoccerChat model achieves the highest weighted F1-score, Cohen’s Kappa, and Matthews Correlation Coefficient (MCC), outperforming alternative approaches such as Q2VL and X-VARS.

For the six-class classification problem, Table~\ref{tab:classification_report_six_class} presents the classification report of the SoccerChat model. The results indicate that "Throw-in" and "Foul" classes achieve relatively high performance, whereas "Goal" and "Shots off target" remain more challenging. The overall accuracy for this task is 57%

These results collectively provide a comprehensive evaluation of the models across different classification tasks, highlighting both their strengths and areas requiring further improvements.

\end{document}